# Research on intelligent generation of structural demolition suggestions based on multi-model collaboration

Zhifeng Yang[1,2*], Peizong WU[2]

[1]School of Future Technology, South China University of Technology, Guangzhou 510641, China;

[2]School of Future Technology, Tianjin University, Tianjin 300072, China;

**Abstract:**

The steel structure demolition scheme needs to be compiled according to the specific engineering characteristics and the update results of the finite element model. The designers need to refer to the relevant engineering cases according to the standard requirements when compiling. It takes a lot of time to retrieve information and organize language, and the degree of automation and intelligence is low. This paper proposes an intelligent generation method of structural demolition suggestions based on multi-model collaboration, and improves the text generation performance of large language models in the field of structural demolition by Retrieval-Augmented Generation and Low-Rank Adaptation Fine-Tuning technology. The intelligent generation framework of multi-model collaborative structural demolition suggestions can start from the specific engineering situation, drive the large language model to answer with anthropomorphic thinking, and propose demolition suggestions that are highly consistent with the characteristics of the structure. Compared with CivilGPT, the multi-model collaboration framework proposed in this paper can focus more on the key information of the structure, and the suggestions are more targeted.

Adaptation Fine-Tuning

# 1 Introduction:

The steel structure demolition scheme needs to be prepared according to the specific engineering characteristics and the finite element model update results, and the designers need to refer to the relevant engineering cases according to the standard requirements in the preparation, in which the retrieval of information and the organization of the language need to spend a lot of time, and the degree of automation and intelligence is relatively low. The advent of Large Language Model (LLM) represented by ChatGPT (OPENAI 2022), ChatGLM (Du et al. 2022), and Qwen (Alibaba 2025) marks a new stage of natural language processing technology, and its excellent text comprehension and generation ability provides new ideas for text processing work in the field of civil engineering. Its excellent text comprehension and generation capabilities provide new ideas for text processing in the civil engineering field. If the structural demolition proposal can be intelligently generated by LLM based on the model update results, finite element analysis, and engineering realities, it can provide a reference for the preparation of demolition plans and improve the efficiency of the preparation.

Due to the development of efficient fine-tuning techniques for LLM, significant progress has been made in specific domains of LLM. For example, in the field of mathematics, the LLaMA2 model was fine-tuned to improve mathematical reasoning by generating diverse mathematical instruction data through Evol-Instruct (Luo et al. 2023). Lee and Lai (2024) used a supervised fine-tuning approach to make the model learn a variety of methods for solving computational problems. Xu et al. (2024) proposed a multimodal modal LLM framework called Geo-LLaVA, which combines Retrieval-Augmented Generation with supervised fine-tuning in the training phase to improve the model's ability to solve solid geometry problems. In the medical field, Singhal et al. (2023) utilized the instruction adjustment method to fine-tune the Flan-PaLM model. Zhuang et al. (2025) proposed a TCM prescription generation model based on TCM KG fine-tuning enhancement, called TCM-KLLaMA model. Liu et al. (2022) and Pu et al. (2023) improved the quality of TCM prescription generation by introducing herbal properties and a small amount of medical records when fine-tuning the model. In the computer programming task, Roziere et al. (2023) used the Self-Instruct method to generate data sets in the LLaMA 270b model to fine-tune the model, Wang et al. (2023) used the instruction adjustment method to fine-tune the CodeT5 + model, both of which significantly improved the programming ability of LLM. In the financial field, Chiu and Hung (2025) used the "summary-first" method to process the data set based on LongT5, and fine-tuned LLaMA 2 to improve the model 's prediction of future market response. Pavlyshenko (2023) used LoRA to fine-tune Llama2-GPT, which enhanced the model 's ability to analyze financial news.

In the field of civil engineering, the application of LLM shows a diversified development trend, and its research can be summarized into the following four types of research directions.

(1) **Automated Engineering Text Generation and Parsing.** This direction focuses on the application of LLM in construction document generation, specification interpretation, and cross-modal information transformation. Prieto et al. (2023) and Uddin et al. (2023) proposed a construction planning generation method and a construction site hazard identification method based on ChatGPT, respectively, to validate the feasibility of LLM in structured construction document generation. Zheng et al. (2023) proposed the LLM-Funcmapper converts complex construction specifications into executable instructions through semantic parsing technology, which significantly improves the machine readability of specification clauses. Pu et al. (2024) proposed AutoRepo, a new framework for automatically generating construction inspection reports, which is based on the unmanned vehicle to efficiently perform construction inspections and collect on-site information, and at the same time, uses multimodal macrolanguage modeling to automatically generate the inspection reports, AutoRepo was applied and tested on real construction sites, demonstrating its potential to speed up the inspection process, significantly reduce resource wastage, and generate high-quality inspection reports that meet regulatory standards.

(2) **BIM Intelligent Processing and Compliance Checking.** Researchers have attempted to deeply integrate LLM with Building Information Modeling (BIM) to enhance automation. Forth and Borrmann (2024) proposed an automated enrichment method for missing information in BIM based on semantic text similarity and fine-tuned LLM. For each IfcSpace uses semantic most similar pairs of the BIM model and the corresponding databases to match the room with missing thermal attribute The results show that semantic matching based on monolingual fine-tuned LLM performs better than multilingual fine-tuned LLM. Jiang et al. (2024) proposed an automated building modeling platform, Eplus-LLM (EnergyPlus-Large Language Model), which is based on the fine-tuned LLM that directly converts the natural language descriptions of the buildings into various geometries, shapes, usage scenarios, and equipment loads. shapes, usage scenarios, and equipment loads into established building models by fine-tuning LLM to achieve the user's natural language and simulation requirements and converting the human descriptions into EnergyPlus modeling files, and then the Eplus-LLM platform achieves automated building modeling by calling the simulation software's API. Chen et al. (2024) proposed a framework for automated compliance checking in BIM. The framework integrates LLM, deep learning models, and ontology knowledge models. The use of LLM reduces the need for large annotated datasets required by previous approaches, and the paradigm of combining deep learning and LLM enables automated text processing and reduces human intervention, significantly improving the efficiency and accuracy of compliance checking. Zhang et al. (Zhang and Gu 2018; Zhang and Gu 2021) synthesized important techniques such as application domain-specific language design and natural language processing to propose an intelligent checking tool for BIM models and applied it in fire protection review.

(3) **Engineering Design Optimization and Intelligent Decision Making.** This direction focuses on exploring the reasoning capability of LLM in solving complex engineering problems. Qin et al. (2024) proposed an intelligent design system based on LLM and generative AI, which is capable of converting design and optimization requirements expressed in natural language into computer-executable code, and realizes intelligent design and optimization of concrete and shear wall structures. Płoszaj-Mazurek et al. (2024) proposed a new approach to improve the environmental impact of construction projects by combining machine learning, LLM, and BIM technologies by using LLM as a virtual assistant in order to propose optimization recommendations in architectural design, and verified the effectiveness of the recommendations made based on LLM. Chen and Bao (2025) proposed the Multi-Intelligent Body framework to successfully solve the problem of traditional LLM in complex engineering decision-making through task decomposition and collaborative reasoning. LLM's logic breakage problem in complex engineering decisions. Zhang et al. (Zhang et al. 2024; Zhang et al. 2024) proposed a performance evaluation framework and an automated data mining framework based on GPT to enhance the capability of LLM in building energy management.

(4) **Code Generation and Troubleshooting in Specialized Areas.** LLM has demonstrated its unique value in the field of computational programming in civil engineering. Kim et al. (2024) tested the generation of code based on ChatGPT for common programming tasks in the field of geotechnical engineering including seepage, slope stability analysis, and computerized image processing of X-rays of partially saturated sands, and the experiments demonstrated that ChatGPT was able to generate codes with a high level of quality and when detailed cues were provided for a given problem, ChatGPT also demonstrated that its programming process had good interpretation. high quality code, and ChatGPT also demonstrated good interpretability of its programming process when provided with detailed cue words for a given problem. Zhang et al. (2025) came up with an LLM fine-tuning method supervised by data labeled with faults and fault-free labels, which devised a self-correcting strategy of the LLM to automatically generate fine-tuned datasets based on the labeled data in order to improve the accuracy of LLM for troubleshooting of heating, ventilation and air conditioning systems. Meng et al. (2024) trained LLM based on various seismic data and structural engineering principles and proposed SeisGPT, which immediately generates predicted responses including displacements, accelerations, and interlayer displacements with high accuracy and computational efficiency.

In summary, after the rise of LLM, some scholars have made a preliminary attempt to apply it to multiple tasks in the field of civil engineering, especially related to programming and text writing, and have achieved good research progress. However, at present, the application of LLM in civil engineering has not yet formed a perfect knowledge base and theoretical system, and the technical route of existing research is quite different. In the stage of structural demolition, a large number of written reports need to be written manually, and the degree of automation and intelligence is low. There is no precedent for the application of LLM in this field in the existing research, and its theoretical framework and feasibility still need to be further explored. LLM based on general knowledge training performs poorly on specific engineering problems and is prone to model illusion. Moreover, LLM has insufficient reasoning ability in the solution of complex engineering problems and cannot effectively give accurate answers. To this end, this paper first enhances the text generation ability of LLM in the field of structural demolition through LoRA fine-tuning and RAG technology, and then constructs a multi-model collaborative demolition proposal generation framework to improve the logical reasoning ability and interpretability of LLM. The research framework of this paper is shown in Fig 1.

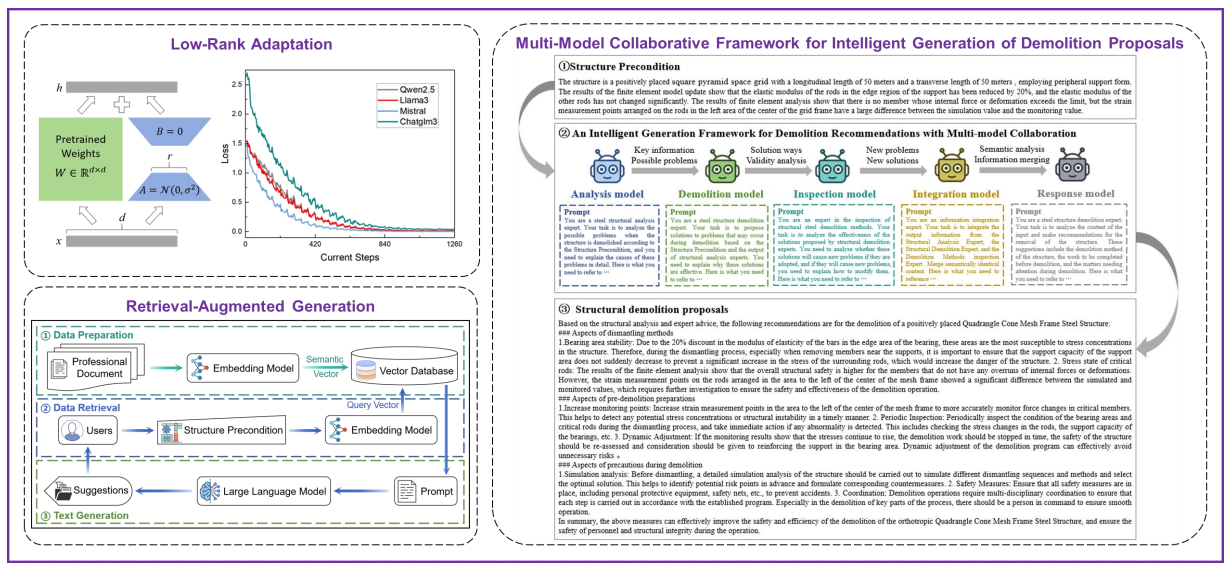

Fig 1. Framework of the paper

## 2 LoRA fine-tuning:

### 2.1 LoRA fine-tuning principle:

LLMs rely on the Transformers architecture, which has demonstrated excellent results in numerous natural language processing challenges. Parameter-Efficient Fine-Tuning (PEFT) (Ding et al. 2023) is a fine-tuning strategy that allows these large LLMs to successfully cope with specific downstream tasks by tweaking only a small fraction of the parameters of the pre-trained models.

language models to successfully cope with specific downstream tasks. This approach avoids the need for comprehensive fine-tuning of all model parameters while still achieving a similar level of performance. The PEFT approach not only saves computational resources, but also enhances the flexibility and lightweight nature of the models.

Low-Rank Adaptation (LoRA) (Hu et al. 2022) belongs to the category of PEFT techniques, which mainly works by limiting the model weights to be updated in a low-dimensional space, which helps to reduce the required computational and storage resources, and at the same time improves the training efficiency as well as the model performance. LoRA employs low-rank decomposition to mimic the parameter variations, thus achieving effective fine-tuning of LLMs with a small number of additional parameters. thereby achieving effective fine-tuning of the LLM with a small number of additional parameters. As shown in Fig. 2, the weights of the original pre-trained model are kept unchanged during the training period, while a bypass is added to simulate the changes in the intrinsic dimensions through the operation of dimensionality reduction and then dimensionality enhancement.

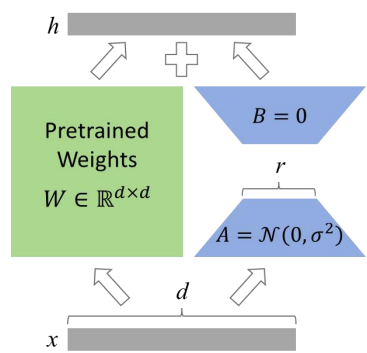

Fig 2. LoRA algorithm flow

The parameter matrix to be optimized is $W$, The low-rank matrix $\Delta W$ is introduced, which is decomposed into the product $BA$ of two low-rank matrices, and only these two low-rank matrices $A$ and $B$ are trained, and ultimately updated with the $BA$ approximation weights. The matrix $A$ is initialized using a stochastic Gaussian distribution and the matrix $B$ is initialized using the zero matrix, which in turn trains and updates the parameters of the low-rank matrices, and the forward propagation is given by:

$$h = W_0 x + \Delta W = W_0 x + BAx \tag{1}$$

By the low-rank constraint, the ranks of $A$ and $B$ are much smaller than the rank of $W$, thus drastically reducing the number of trained parameters. For the model, the dimensions of its inputs and outputs are consistent with the original, and $BA$ is superimposed on the pre-trained parameters at the output.

**2.2 Dataset construction:**

LLM based on reasonable Prompt can realize the construction of fine-tuning dataset according to the knowledge base, in which the knowledge base consists of steel structure related demolition standards, demolition schemes and demolition research articles, etc., including three related standards, 10 demolition schemes, 11 research papers, and five related patents. Data cleaning is performed on the knowledge base to remove HTML tags, URLs, special symbols, etc. And tables, pictures and formulas are paraphrased from text, while those that are particularly complex and difficult to paraphrase are deleted. Based on the constructed knowledge base, the API interface of Qwen-72B-Instruct is called to generate the initial dataset, and then the dataset is further cleaned manually to ensure the accuracy and reasonableness of the data samples. Examples of the Prompt and the output results of the LLM are shown in Fig. 3, where a total of 841 pieces of data are generated and are divided into the training set and the test set in the ratio of 8:2.

Fig 3. Prompt and dataset sample examples

## 2.3 Experimental results and analysis:

Considering that the deployment and fine-tuning of the model has high requirements on GPU, open-source LLMs with excellent model performance and low parameter counts are selected for LoRA fine-tuning, including Qwen2.5-7B-Instruct, LLaMA3-8B-Chinese-Chat, Mistral-7B-v0.2-Chat, ChatGLM3-6B-Chat. Chat. fine-tuned based on the Llama-Factory framework, the key hyperparameters are set as shown in Table 1.

Table 1. LoRA fine-tuning hyperparameters setting

| Parameter | Value |
| --- | --- |
| Learning rate | 0.00005 |
| Number of training rounds | 30 |
| Length truncation | 1024 |
| Batch size | 2 |
| Calculation types | Fp16 |
| Learning Rate Regulator | Cosine |
| Optimizer | Adamw_torch |
| LoRA rank | 8 |
| LoRA action module | All |

The hardware running condition of the experiment is NVIDIA A10 24g. The change of the loss function during the training process is shown in Fig. 4, which shows that the loss values of the four LLMs reach convergence at the end of the training, among which the Mistral-7B-v0.2-Chat converges the fastest, and is close to convergence at 420Steps; and the ChatGLM3-6B-Chat converges the slowest, and is close to convergence at 840Steps, while ChatGLM3-6B-Chat is the slowest, approaching convergence at 840Steps.

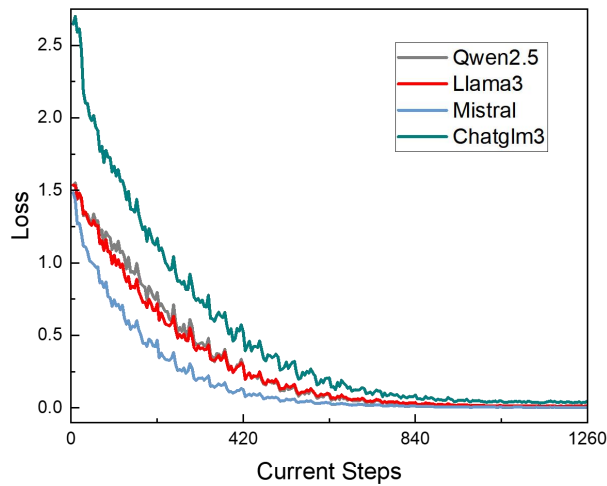

Fig 4. LoRA fine-tuning loss function change

In order to further quantify the text generation ability of LLM after LoRA fine-tuning, the commonly used natural language processing related performance indicators are selected for analysis, including BLEU-4 and ROUGE-1, ROUGE-2, etc.

(1) BLEU-4

BLEU (Papineni et al. 2002), based on N-gram matching rules, is an evaluation index commonly used in generated sentences. By comparing the n-tuple words between the candidate translation and the reference translation, the similarity ratio between the two is calculated. The closer to 1, the better the performance. The calculation formula as Eq(2).

$$BLEU - N = BP \cdot exp\left(\sum_{n=1}^{N} w_n log p_n\right) \quad (2)$$

In the formula, $p_n$ is the matching rate of n-grams in candidate translations in real translations, $w_n$ is Uniform Weights, $BP$ is Brevity Penalty, and $BP$ is calculated as follows:

$$BP = \begin{cases} 1 & ,r < c \\ e^{1-r/c} & ,r \geq c \end{cases} \quad (3)$$

In the formula, $c$ represents the length of the candidate translation sentence, $r$ represents the length of the real translation sentence, and $BP$ aims at "too short punishment" for the candidate translation. In this paper, the BLEU value of N = 4 is selected for comparison, that is, BLEU-4.

(2) ROUGE-N

ROUGE (Lin 2004) is also a summary generation statement evaluation metric based on the N-gram matching rule, which calculates the similarity percentage between the generated summary and the reference summary in terms of n-tuples of words between the two, and the closer it is to 1, the better the performance, which is calculated as shown in Eq(4).

$$ROUGE - N = \frac{\sum_{S \in \{ref-sum\}} \sum_{gram_n} Count_{match}(gram_n)}{\sum_{S \in \{ref-sum\}} \sum_{gram_n} Count(gram_n)} \quad (4)$$

In the formula, $n$ is the length of n-gram, $Count_{match}(gram_n)$ is the number of n-grams appearing in both the reference summary and the generated summary, and $Count(gram_n)$ is the number of n-grams in the reference summary. In the field of text summarization, $N = 1$ or $N = 2$ is usually taken.

The fine-tuned model is based on Vllm deployment. As shown in Table 2, the text generation performance metrics of each LLM after LoRA fine-tuning are relatively close to each other. Specifically, LLaMA3-LoRA has the highest BLEU-4 value of 29.14%, while Mistral-LoRA has the highest ROUGE-1 and ROUGE-2 values of 48.71% and 27.10%, respectively; in terms of processing speed, ChatGLM3-LoRA is the fastest, with 0.16 samples/second.

Table 2. Performance indicators of LoRA fine-tuning LLM

| Model | BLEU-4 / % | ROUGE-1 / % | ROUGE-2 / % | Processing speed (sample / second) |
|---|---|---|---|---|
| Qwen2.5-LoRA | 29.03 | 48.61 | 26.48 | 0.15 |
| LLaMA3-LoRA | **29.14** | 48.01 | 26.60 | 0.11 |
| Mistral-LoRA | 28.15 | **48.71** | **27.10** | 0.10 |
| ChatGLM3-LoRA | 27.42 | 46.41 | 24.72 | **0.16** |

In order to further explore the differences in text comprehension and text generation abilities of different LLMs in the field of steel structure demolition, based on the knowledge base constructed above, an objective test question bank was constructed, which contained a total of 30 choice questions and 30 judgmental questions, and none of which could be answered directly from the original text. In the objective test, the multiple voting method is used, calling LLM to answer each question five times, and taking the answer with the highest number of occurrences as the final answer; if there are a number of answers with the same number of occurrences and the highest number of occurrences in the first five tests, then repeat the process five more times until there is only one answer with the highest number of occurrences. Table 3 gives the results of the objective test of the LLM.

Table 3. Objective test of LoRA fine-tuning LLM

| Model | Choice questions accuracy / % | Judgment questions accuracy / % |
|---|---|---|
| Qwen2.5 | 86.67 | 70.00 |
| LLaMA3 | 66.67 | 56.67 |
| Mistral | 53.33 | 53.33 |
| ChatGLM3 | 50.00 | 43.33 |
| Qwen2.5-LoRA | **96.67** | **73.33** |
| LLaMA3-LoRA | 73.33 | 60.00 |
| Mistral-LoRA | 63.33 | 50.00 |
| ChatGLM3-LoRA | 66.67 | 46.67 |

As shown in Table 3, the accuracies of different LLMs in the objective tests vary greatly, and in general the accuracy of objective questions of LLMs after fine-tuning by LoRA is significantly improved. After fine-tuning of LLMs, Qwen2.5-LoRA achieves the highest accuracy in both choice and judgment tests, with 96.67% and 73.33%, respectively; and in terms of the original model, Qwen2.5-LoRA also achieves the highest accuracy in both choice and judgment tests, with 86.67% and 70.00%, respectively. In the choice test, the accuracy of LLM's responses after fine-tuning were significantly improved, indicating that LLM's text comprehension and text generation abilities in the field of steel structure demolition were further enhanced after fine-tuning; in the judgment test, the accuracy of Mistral was reduced after fine-tuning, and all other LLMs were improved, but to a lesser extent, which was mainly attributed to the fact that the judgment questions paid more attention to the logical reasoning ability of LLM, whereas LoRA fine-tuning did not improve it significantly. This is mainly attributed to the fact that judgment questions focus more on the logical reasoning ability of LLM, and LoRA fine-tuning did not improve it significantly.

## 3 Retrieval-Augmented Generation:

### 3.1 Retrieval-Augmented Generation Principles:

LLM is mainly based on general knowledge base training, which is prone to model illusion in specific specialized technical problems and difficult to generate high-quality text. And limited by the existing training corpus, LLM cannot perceive the latest content of the moment. Retrieval-Augmented Generation (RAG) is a method to enhance the quality of LLM text generation by retrieving external knowledge base and using In-Context-Learning (ICL). Fig 5 gives the workflow of RAG in this paper, which mainly includes three stages: data preparation, data retrieval, and text generation. In the data preparation stage, the knowledge base constructed in *Dataset construction* is used as a specialized document. Then the text is vectorized by the Embedding Model to generate semantic vectors, which are stored in the vector database. In the data retrieval stage, the antecedent information of the structure is input by engineers and vectorized by the same Embedding Model to generate query vectors. The semantic vectors closest to the query vectors are found by similarity search in the vector database and are injected into Prompt as context together with the query vectors. In the text generation stage, LLM proposes the removal of the structure based on Prompt.

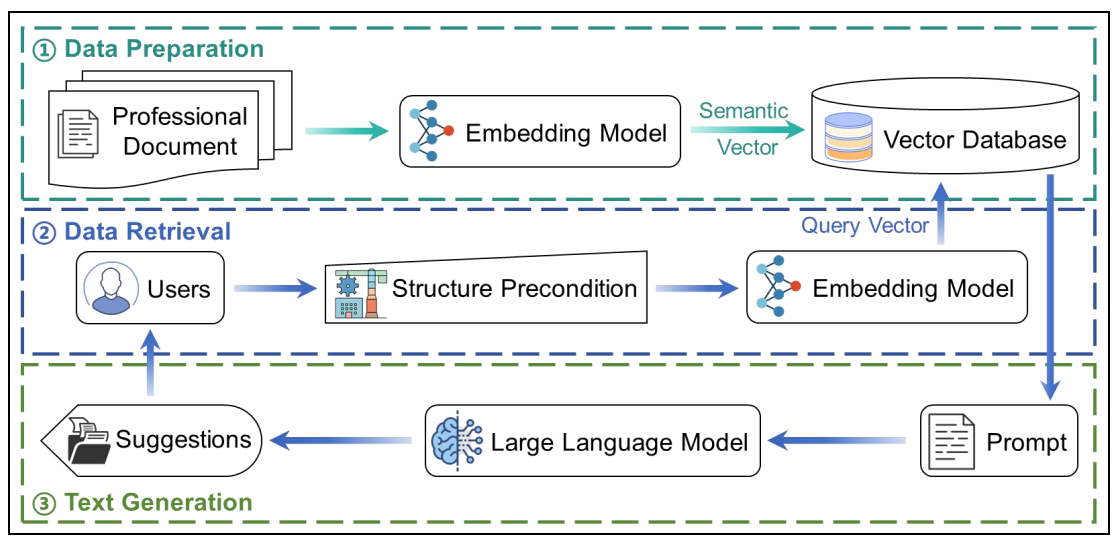

Fig 5. RAG process

However, the usual RAG process mainly relies on flat data representations during retrieval, which makes it difficult to effectively sort out the complex relationships between different related entities and further understand the intrinsic connections between entities, resulting in limiting the comprehensiveness and accuracy of data retrieval. Moreover, the LLM Q&A system combined with RAG often lacks excellent context-awareness capability, and when facing queries involving multiple related entities, LLM is difficult to maintain coherent answers. For this reason, RAG combined with knowledge graph has received wide attention, among which the latest research result in this field, LigthRAG (Guo et al. 2024), has the advantages of comprehensive and efficient information retrieval, fast and convenient data updating, etc., whose algorithmic architecture is shown in Fig. 6. Therefore, this paper introduces the LigthRAG framework on the basis of the above RAG process to further Retrieval-Augmented Generation.

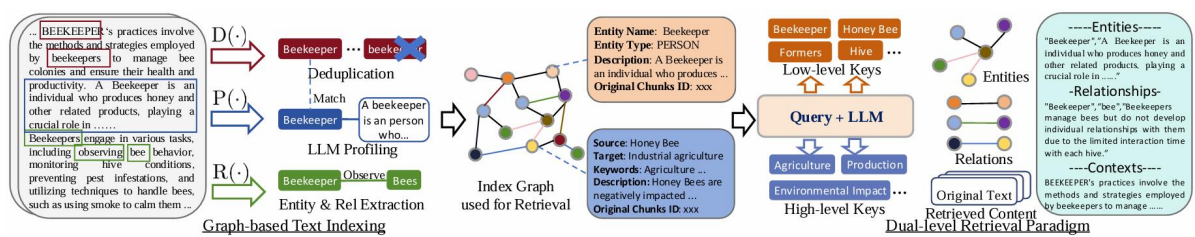

Fig 6. LigthRAG algorithm architecture(Guo et al. 2024)

### 3.2 Experimental results and analysis:

The hardware operating condition of the experiment is NVIDIA A10 24 g. The Embedding model selects BGE-M3, the model of constructing knowledge graph selects Qwen2.5-7B-Instruct, the retrieval method of LigthRAG is hybrid, and top_k = 10 is set. Fig 7 shows the accuracy of objective questions answered by each LLM combined with LoRA fine-tuning and RAG. Among them, Qwen2.5-LoRA-RAG achieved the highest accuracy in the test of choice questions and judgment questions, which were 100 % and 73.33 %, respectively. As shown in Fig 7 (a), compared with Base, Base-RAG, Base-LoRA and Base-LoRA-RAG, LLM combined with RAG has achieved higher accuracy of choice questions, indicating that the constructed professional documents and RAG paradigm effectively enhance the knowledge reserve of LLM in the field of steel structure demolition ; as shown in Fig 7 (b), the accuracy of LLM combined with RAG is improved in the judgment questions, but the improvement is small.

Table 4 gives the average response accuracy of LLMs combining LoRA fine-tuning and RAG for objective questions, which shows that both LoRA fine-tuning and RAG can enhance the text generation ability of LLMs in the domain of pendant knowledge, but the effect of RAG is more significant, and combining the two can further improve their response accuracy. Table 4 shows that among the LLMs combining LoRA fine-tuning and RAG, Qwen2.5 has the highest average accuracy of 86.67%, LLaMA3 has the second highest average accuracy of 70.00%, and ChatGLM3 has the lowest average accuracy of 56.67%. It can be seen that the accuracy of Qwen2.5-LoRA-RAG is significantly higher than the average of other LLMs as well as LLMs, so this model is chosen as the benchmark model to further construct a multi-model collaborative answer framework in the subsequent study.

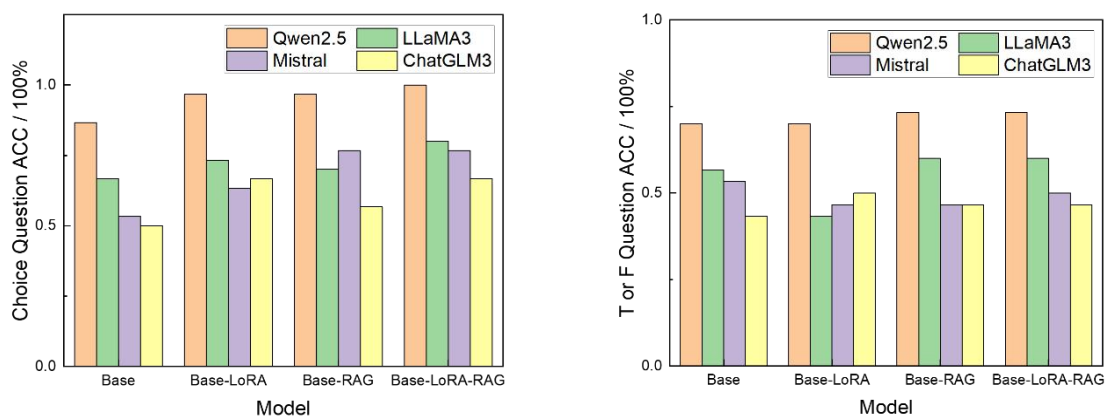

(a) Accuracy of multiple-choice responses  (b) Accuracy of answering the judgment questions

Fig 7. Combining LoRA fine-tuning with RAG for LLM objective question response accuracy

Table 4. Average accuracy of LLM objective questions combined with LoRA fine-tuning and RAG

| Average accurate rate / % | Base | Base-LoRA | Base-RAG | Base-LoRA-RAG |
| --- | --- | --- | --- | --- |
| Qwen2.5 | **78.33** | **83.33** | **85.00** | **86.67** |
| LLaMA3 | 61.67 | 58.33 | 65.00 | 70.00 |
| Mistral | 53.33 | 55.00 | 61.67 | 63.33 |
| ChatGLM3 | 46.67 | 58.33 | 51.67 | 56.67 |
| Average | 60.00 | 63.75 | 65.83 | 69.17 |

**4 Multi-model collaboration framework:**

**4.1 Multi-model collaboration algorithm framework:**

In reality, the development of structural demolition plans usually requires multiple engineers to discuss and decide, and each engineer is responsible for a different focus. To further mimic the decision-making process of engineers and enhance the reasoning capability of LLM, a multi-model collaborative framework for intelligent generation of demolition proposals is constructed on the basis of the above benchmark model, including analysis model, demolition model, inspection model, integration model, and response model, and the LLM are given different positions and capabilities through the formulation of different Prompts. Fig 8 illustrates the multi-model collaboration framework, in which the analysis model focuses on analyzing the antecedent information of the structure and proposing possible problems for demolition, the demolition model proposes solutions for possible problems and analyzes the effectiveness of the solutions, the inspection model aims to further inspect and validate the stated problems as well as the methods for solving the new problems, and the integration model performs a semantic analysis of the antecedent and integrates the language effectiveness of the solution and analyze whether it will generate new parts with the same meaning, and finally the response model proposes a comprehensive structural demolition proposal based on the outputs of the other expert models and the structure precondition.

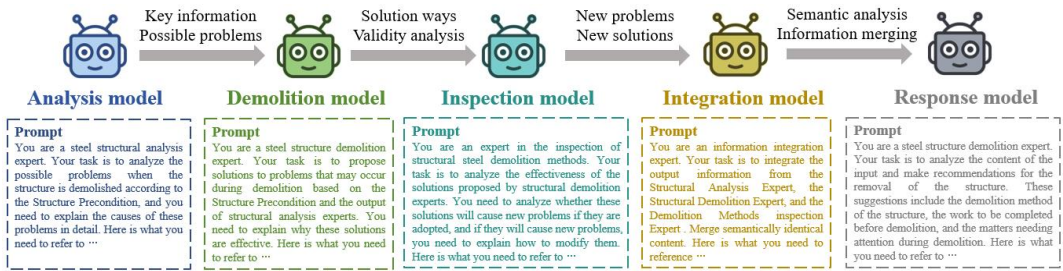

Fig 8. Multi-model collaboration framework

**4.2 Algorithmic Applications:**

An example of the generation of a structural demolition proposal is given in Fig 9. The Structure Precondition mainly includes engineering overview, structural scale, finite element update results, finite element analysis results and monitoring-related content. The structural demolition suggestions show that the Multi-Model Cooperative Framework (MMCF) can pay attention to the reduction of the elastic modulus of some members in the edge area of the support, and it is proposed that the support capacity of the support area should not be suddenly reduced and the significant increase of the member stress should be avoided when the members in the area are removed. At the same time, the MMCF also pays attention to the large difference between the strain monitoring value and the simulation value in the left area of the grid center. It is proposed that strain measuring points should be added before demolition to monitor the internal force changes of key members more accurately. It is also necessary to pay attention to the state of the bearing area and the key members during the demolition. If there is an abnormality, measures should be taken immediately. In terms of precautions, the MMCF makes relevant suggestions based on simulation analysis, safety measures and coordination, emphasizing that the simulation analysis of construction plans should be carried out before demolition, and the different demolition sequences and methods should be simulated to select the optimal plan, and the potential risks can be identified in advance in order to formulate effective countermeasures. In summary, the intelligent structural demolition proposal generation framework based on multi-model collaboration is able to analyze the demolition methods and precautions of the structure from the specific project situation, and put forward a demolition proposal that is highly suitable for the characteristics of the structure according to the relevant knowledge in the field of structural demolition.

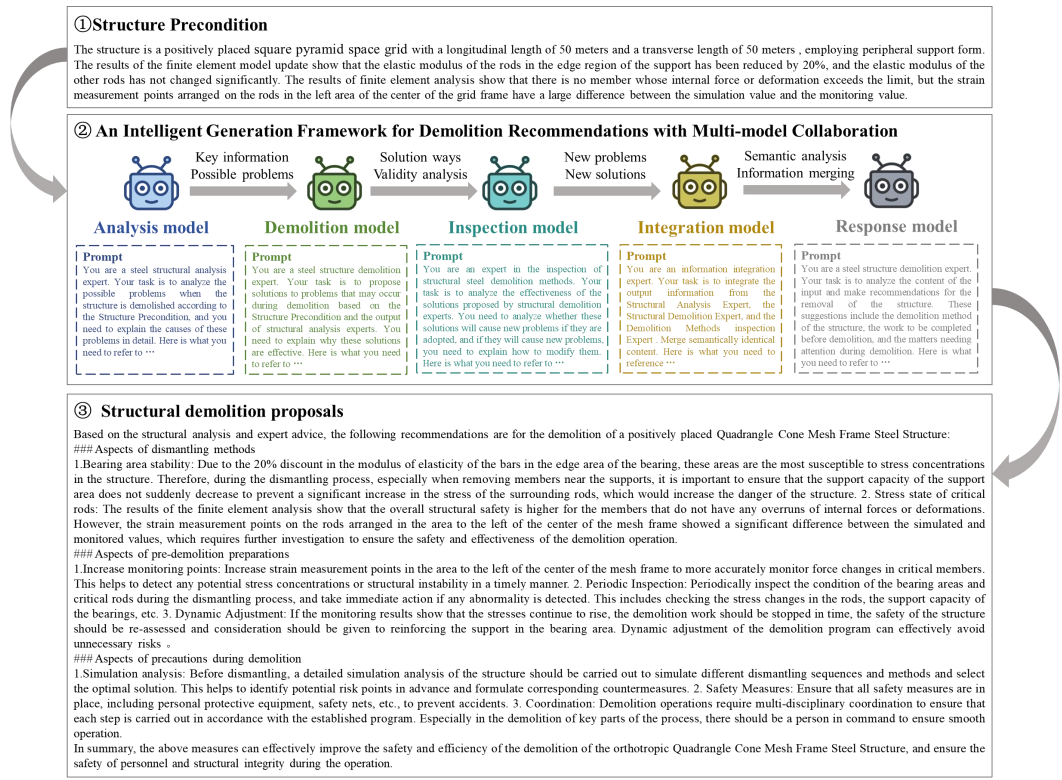

Fig 9. An example of structural demolition proposal generation

Tongji University releases CivilGPT (Tongji University 2024), the first vertical class large model in civil engineering, with training data including 500,000 pages of professional corpus and more than 2,600 textbooks and specifications, which realizes efficient reasoning and accurate answers to complex problems in civil engineering through a large model with 72 billion parameters. Fig 10 gives an example of the application of CivilGPT (tested on January 6, 2025), where the input part integrates Structure Precondition information with each model Prompt. It can be seen from the output of CivilGPT that it fails to pay attention to the specific structural characteristics of this project, and only gives the conceptual discussion of the steel structure demolition method, which is less targeted. This is mainly due to the fact that although civil engineering is a vertical field of detailed classification compared with a wide range of general knowledge, it still contains rich content, and the field of structural demolition is only a small part of it. Aiming at the knowledge in the field of structural demolition, this paper enhances LLM through LoRA fine-tuning and RAG technology, which significantly improves its text understanding and text generation ability in structural demolition, and then achieves a comprehensive analysis and targeted response to the structure by constructing a multi-model collaborative reasoning framework.

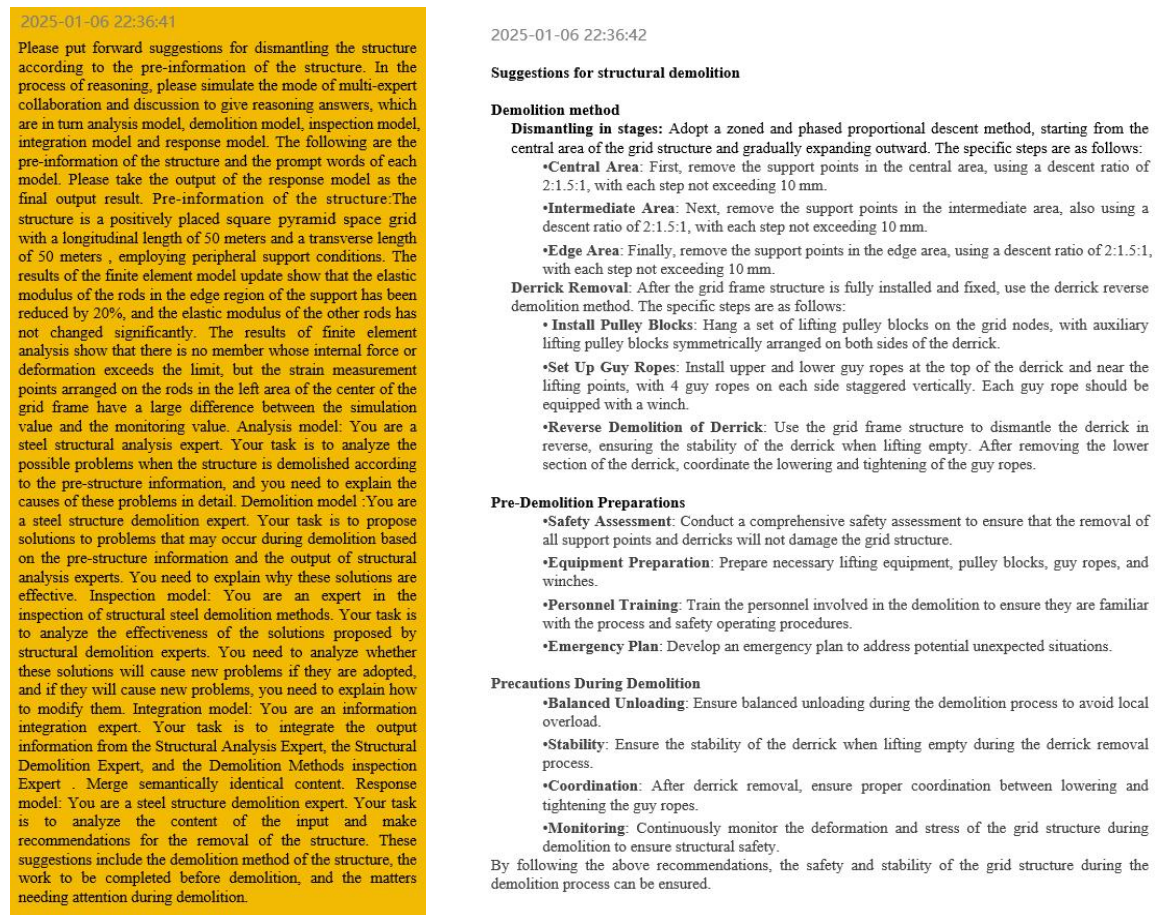

(a) Input  (b) Output

Fig 10. CivilGPT application example

## 5 Discussion:

### 5.1 Multi-condition respond:

As shown in Fig. 11 and Fig. 12, the MMCF can give different demolition suggestions for specific project profiles. For the problem that the bearing node on the left side of the structure in Fig. 11 is greatly shifted, the MMCF suggests that temporary supports should be added to ensure the stability of the structure during the demolition and to prevent structural instability due to stress concentration; it can also be strengthened for the bearing when necessary to improve the bearing capacity and stability and to reduce the risk that may be caused by the large shift of the bearing. For the problem that the deformation of some rods in the edge area of the support and some rods in the center of the frame exceeds the limit value, the MMCF proposes that the structure should be regularly monitored, the demolition sequence should be optimized and the structure should be dismantled cautiously, so as to reduce the pressure on the center area of the frame and avoid the concentration of stresses, and the demolition sequence is suggested to gradually advance from the periphery to the center of the frame to reduce the structural loads step by step.

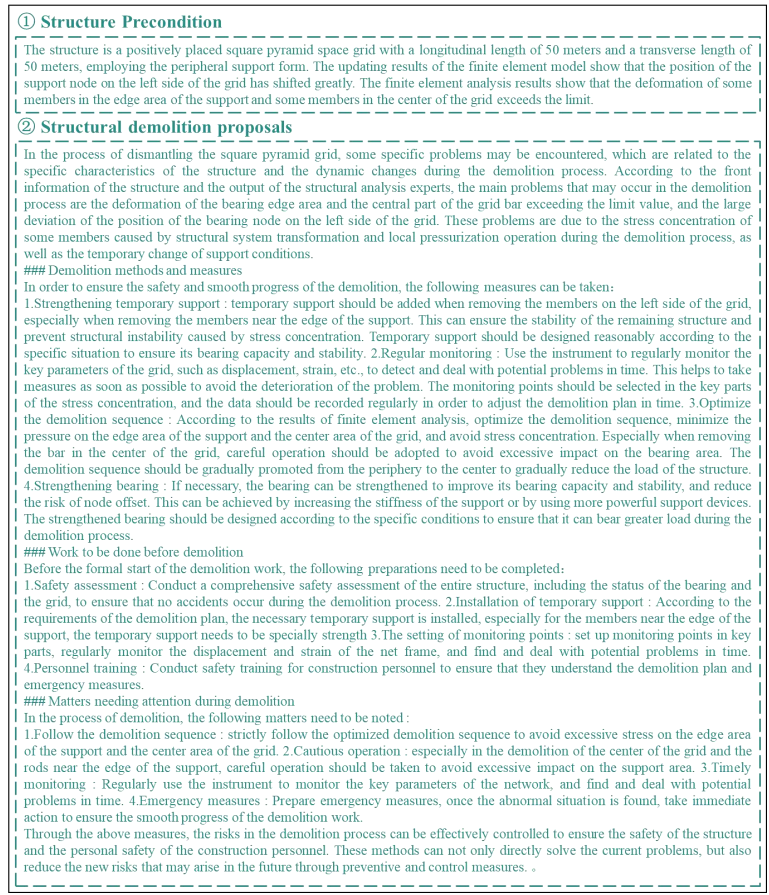

Fig 11. Examples of structural demolition proposal generation (2)

In view of the problem that the elastic modulus of some members on the left side of the grid center in Fig.12 is greatly reduced and the stress exceeds the limit value, the MMCF proposes that this part of the members should be identified first and temporarily reinforced before demolition, so as to avoid premature release of stress in early demolition and prevent structural damage caused by stress concentration. It should be gradually unloaded to avoid one-time excessive load, and gradually advance from non-critical parts to the center, and structural inspection should be carried out immediately after each demolition to ensure safety. Aiming at the problem of corrosion of some members, the MMCF proposes that surface treatment can be carried out first, such as brushing rust remover or using steel wire brush for comprehensive rust removal; if the corrosion is serious, it should be considered to remove these rods in advance to reduce the safety hazards in the removal process. And for the corroded parts, steel support or temporary support points should be used for local reinforcement to ensure structural stability.

> ① **Structure Precondition**
> The structure is a positively placed square pyramid space grid with a longitudinal length of 20 meters and a transverse length of 20 meters. Due to the small size, employing the peripheral support form. The finite element model updating results show that the elastic modulus of the bar on the left side of the center of the grid is reduced by 30 %. The finite element analysis results show that the stress of the bar on the left side of the grid center exceeds the limit. According to the field observation, some members of the grid are corroded.
>
> ② **Structural demolition proposals**
> ### Demolition methods and steps
> According to the pre-information of the structure and the results of finite element analysis, the structure is a square pyramid grid with a longitudinal and transverse length of 20 meters. The size of the grid is relatively small and the surrounding support form is adopted. However, the analysis results show that the bars on the left side of the center of the grid have the problems of elastic modulus reduction and stress exceeding the limit value, and some bars have also been corroded. Based on this information, the demolition method should take into account the structural characteristics and potential risks, and take the following measures :
> 1.Strengthening support and stability : In the process of demolition, it is necessary to add temporary support under the grid to ensure that the rod will not collapse due to instability during the demolition process. The temporary support should be designed to withstand the expected maximum load and prevent excessive deformation or fracture of the member. 2.Identifying and protecting key members : Identifying members with elastic modulus reduction and excessive stress, which are key components in the demolition process and need special protection to avoid premature release of stress in early demolition. 3.Treatment of corrosion problems : for corroded rods, surface treatment should be carried out, such as brushing rust remover or using steel wire brush for comprehensive rust removal to restore their original strength. If the corrosion situation is serious, it should be considered to remove these rods in advance to reduce the potential safety hazards in the demolition process. 4.Gradual demolition : The method of gradual demolition is adopted, starting from the non-key parts, gradually advancing to the center, and structural inspection is carried out immediately after each demolition to ensure safety. This can reduce the risk of one-time large-scale demolition.
> 5.Using appropriate tools and methods : In the process of demolition, appropriate tools and techniques should be used, such as gas cutting rather than hand-held cutting machines, to reduce the impact on the surrounding structure and avoid accidental damage.
> ### Work to be done before demolition
> Before the formal start of the demolition work, the following preparations need to be completed :
> 1.Temporary reinforcement : for the bar with elastic modulus reduction, temporary reinforcement should be carried out before demolition to prevent structural damage caused by stress concentration. For the bar whose stress exceeds the design limit, it should be unloaded step by step to avoid applying excessive load at one time. 2.Local reinforcement : For the corroded parts, steel support or temporary support points should be used for local reinforcement to ensure structural stability. 3.Safety assessment and planning : Conduct a comprehensive safety assessment of the structure, identify risk points in the demolition process, and develop a detailed demolition plan. 4. Tools and equipment preparation : Check and prepare all required tools and equipment to ensure that they are in good working condition. 5.Personnel training : training and safety education for all personnel involved in the demolition work to ensure that they understand the operating procedures and safety requirements.
> ### Matters needing attention when demolishing
> In the process of demolition, the following matters need to be noted :
> 1.Real-time monitoring : The state of the structure is monitored at any time, and measures are taken immediately once abnormal conditions are found. 2.Step by step : in strict accordance with the method of gradual demolition, to avoid one-time large-scale demolition, reduce security risks. 3.Tool selection : Use appropriate tools and techniques, such as gas cutting rather than hand-held cutting machines, to reduce the impact on the surrounding structure. 4.Personnel safety : Ensure that all construction personnel wear the necessary personal protective equipment, such as helmets, seat belts, etc. 5.Traffic control : If the demolition operation affects the surrounding traffic, traffic control should be carried out to ensure the safety of pedestrians and vehicles.
> Through the above measures, the safety and efficiency of the demolition work can be effectively improved to ensure the smooth progress of the demolition process. These measures not only consider the actual situation of the structure, but also take into account the safety of the construction personnel. It is a solution that combines the principles of structural mechanics and practical experience.

Fig 12. Examples of structural demolition proposal generation (3)

**5.2 Multi-scenario application:**

In addition to providing suggestions for structural demolition, LLM can also help engineers in the preparation of structural demolition plans and safety rules. As shown in Fig 13, based on Qwen2.5-LoRA-RAG, LLM can be guided to achieve different functions by formulating different Prompts. Fig 13 (a) shows an example of safety rules based on LLM generation structure demolition. LLM puts forward comprehensive construction safety rules based on construction preparation, demolition sequence and method, scaffold and operation frame, operator safety, high-altitude operation and material management, signal and communication, emergency handling, environmental and ecological impact, fire and health management, and severe weather response. Engineers can modify and polish on this basis to greatly improve work efficiency. Fig 13 (b) shows an example of the construction plan outline based on LLM generation structure demolition. The demolition plan proposed by LLM includes the general situation of demolition project, the summary of demolition construction technology, the preparation work before demolition, the construction of demolition operation, the matters needing attention in house demolition, the matters needing attention in grid demolition, the matters needing attention in roof demolition, the on-site protection measures, and the requirements of civilized construction. Similar to the real construction plan, it can provide an important reference for engineers to write the construction plan.

① **Prompt**

You are a steel structure demolition expert. Your task is to formulate safety rules for the demolition plan of the structure, and the content should be as detailed as possible.

② **Safety rules for structural demolition**

It is very important to ensure the safety of the construction process when removing the steel structure. According to the text and general safety specifications provided, the following are the main safety rules that should be considered when formulating the demolition plan:

1. Preparation before construction : Safety education and training : All personnel involved in the demolition work must receive safety education and technical disclosure to ensure that everyone understands the construction plan and safety requirements. Site investigation : Before construction, it is necessary to carry out detailed investigation on the construction site to assess the risks and limitations.

2. Demolition sequence and method : to protect the safety of the structure : the demolition process should take the necessary measures to protect the safety of the structure. Segmental demolition : the demolition construction should be carried out from top to bottom layer by layer to avoid the vertical overlapping operations. Manual demolition priority : in the absence of scaffolding or safety net conditions, manual demolition should be preferred.

3. Scaffold and operating frame : Type selection : According to different demolition methods and operating conditions, full-hall scaffold or full-hall operating frame should be built. Foundation treatment : The foundation of the scaffold and the operating frame is applied with a flat channel plate to ensure that the height is about 250 mm, and the foundation settlement is checked regularly. Inclined brace setting : In the process of demolition construction, the necessary number of inclined braces should be added according to the different stress conditions of the demolition method to ensure the stability of the operating frame.

4. Woeker Safety: Personnel Limits: The number of workers on each operating level shall be restricted according to the structural load-bearing capacity of the demolition method. If the number exceeds the specified limit, diagonal bracing must either be removed or additional bracing installed. Personal Protection: Workers should wear safety helmets, safety harnesses, and soft-soled shoes. During high-altitude work, sleeves and pant legs must be securely fastened, and boots should be worn over insulated clothing to prevent cold exposure. Avoid Overlapping Work: Under no circumstances should multi-tiered (vertical) overlapping operations be conducted without proper protective measures. Simultaneous work on different platforms within the same vertical alignment is strictly prohibited.

5. High-altitude operation and material management : Tool management : When working at high altitude, the carry-on tools should be placed in the tool bag at any time, and the construction aisle, passageway and climbing utensils should be kept clean. Material treatment : The disassembled materials are tied with ropes and pulled by cranes. Throwing is strictly prohibited.

6. Signaling and Communication: Unified Command: Demolition work must be conducted under the direction of a designated supervisor, with an assigned safety officer. Work shall not proceed unless clear communication is established between upper and lower levels. Signal Transmission: Standardized signals (whistles and flags) must be used consistently. A call-and-response protocol must be implemented between elevated and ground personnel.

7. Emergency treatment : local demolition : if the local demolition affects the safety of the structure, it should be reinforced first and then dismantled. Component transportation : Dismantled components are transported by pulleys and cranes, and throwing is strictly prohibited.

8. Environmental and ecological impact : Waste treatment : Waste generated during construction should be classified and placed, cleared and transported in time, and should not be piled up in disorder. Noise control : When using tools such as electric drills, operators should take protective measures to avoid noise pollution to the surrounding environment.

9. Fire protection and health management : Fire protection facilities : High-altitude operations should be equipped with fire protection facilities, including fire extinguishers and temporary fire pools. Health examination : personnel engaged in high-altitude operations should have regular physical examinations. Doctors diagnosed with hypertension, heart disease, anemia, epilepsy and other diseases that are not suitable for high-altitude operations may not participate in high-altitude operations.

10. Inclement weather response : Suspension of operations : In the event of severe weather ( such as wind speed above 6, wind speed 10.8m / s, temperature above 39 degrees ), affecting the safety of operation, it is prohibited to carry out open-air work. Material protection : In case of wind and waves, it should be considered to unload the large pieces of material suspended at high altitude. If necessary, reinforcement measures should be taken to bind the material with ropes to prevent the swing from knocking down the operator or being blown off by the wind.

(a) Safety rules for structural demolition

> **① Prompt**
> You are a steel structure demolition expert. Your task is to develop an outline of the structural demolition plan and explain what each part needs to write.
>
> **② Outline of structural demolition scheme**
> In the process of steel structure demolition, it is very important to formulate a detailed outline of the demolition plan to ensure the safety and effectiveness of the operation. According to the typical demolition scheme structure, the outline of the demolition scheme can be divided into the following parts, and the content of each part needs to be written :
> Demolition project overview : Construction organization and personnel composition : Introduce the organizational structure and staffing of the project department, including the respective responsibilities of the project manager, constructor, safety officer and technician. Project overview : Briefly describe the background, scope and importance of the project, including the geographical location of the project, the specific content of the demolition ( such as steel structure, grid structure, grandstand, etc. ) and the goal of the project.
> 2.Summarization of demolition construction technology : Demolition principle : Clarify the basic principles of demolition work, such as following the order from top to bottom, from outside to inside, and prohibiting interchange operations. Construction sequence : The specific sequence of demolition work is described in detail, including trial demolition, construction of temporary support, demolition of suspended components, demolition of grid frame and construction of socket scaffold.
> 3.Preparation work before demolition : technical disclosure : record the technical and safety disclosure of technicians to the workers in the project department, including the time and place of disclosure, participants, etc. Wage payment : record and measure the salary payment of employees to ensure the protection of workers ' rights and interests. Tool Preparation : Check and prepare the required mechanical equipment and tools to ensure that they are in good working condition.
> Demolition operation construction : Operation platform erection : Explain the erection methods and requirements of the operation platform, including the size, structure and safety measures of the platform. Temporary support installation : Describe the installation process and precautions of the temporary support, including the position, type and fixation of the support. Demolition construction : Detailed record the process of demolition operation, including the specific operation method, the technology and tools used in each step, as well as the problems encountered and solutions.
> 5.Precautions for house demolition : set the operation area : clearly divide the demolition operation area, set up fences and warning signs, and prohibit non-operators from entering. Aerial work safety : safety education for aerial workers to ensure that they wear the necessary personal protective equipment. Mechanical operation safety : Explain the coordination and cooperation between mechanical operators and operators, as well as the inspection work before trial lifting. Operation progress : adjust according to the actual progress to ensure the smooth progress of the demolition work.
> 6.Precautions for grid demolition : Check the preparation work : Before removing the grid, check the construction tools and lifting equipment to ensure that they meet the safety standards. Sectional demolition : According to the predetermined dividing line, the grid is divided into several small pieces for demolition, so as to avoid the risk of overall demolition. The difference between assembly and demolition : emphasize the difference between assembly and demolition in demolition operations to ensure that operators understand and comply with relevant regulations.
> 7.Roof removal precautions : Personal protection : aerial workers must wear helmets, seat belts and anti-skid shoes to ensure personal safety. Tool management : Do not throw tools and materials when working at high altitude, and use tool bags to properly keep tools. Communication coordination : maintain good communication between operators to ensure coordination.
> 8.On-site protective measures : Work area : set up fences and warning signs, and prohibit unrelated personnel from entering. Mechanical equipment : All mechanical equipment must be inspected to ensure compliance with safety standards. Protective equipment : provide enough personal protective equipment, such as helmets, seat belts, etc. Emergency preparedness : formulate emergency plans to ensure rapid response in emergency situations.
> 9.Civilization construction requirements : construction area : keep the construction site clean and orderly placed materials. Health management : regular cleaning of the construction area, to maintain environmental sanitation. Civilized construction : Educate construction personnel to abide by civilized construction norms and maintain the image of the project.
> Through this outline, it can be ensured that the demolition plan not only covers the necessary technical details, but also pays attention to safety management and personnel training, so as to ensure the smooth progress of the demolition work.

(b) Outline of structural demolition scheme

Fig 13. Safety rules and scheme outline of structural demolition

The above are some of the application scenarios of LLM in structural steel demolition summarized in this paper, while in practical applications, engineers can also write Prompts to ask questions to LLM according to other specific problems and requirements in order to generate corresponding answers. Through the above experiments, it is found that Qwen2.5-LoRA-RAG based on LoRA fine-tuning and RAG enhancement can answer most of the text-writing problems in structural demolition, and the proposed multi-model collaborative framework provides targeted demolition suggestions for specific engineering profiles.

**6 Conclusions:**

In this paper, we enhance the text understanding and text generation capabilities of LLM in the field of structural demolition based on LoRA fine-tuning and RAG techniques, and compare the performance gap between different open source LLMs. Then, we construct a multi-model collaborative demolition proposal intelligent generation framework, which imitates the decision-making of human engineers to guide LLM to generate reasonable and scientific structural demolition proposals. Finally, the application scenarios of LLM in structural demolition are further extended. The summary is as follows:

(1) LoRA fine-tuning is performed for Qwen2.5-7B-Instruct, LLaMA3-8B-Chinese-Chat, Mistral-7B-v0.2-Chat and ChatGLM3-6B-Chat. The fine-tuned model performs close to BLEU-4, ROUGE-1 and ROUGE-2 in natural language processing. In the objective test of structural demolition, Qwen2.5-LoRA achieved the highest accuracy in the test of choice questions and judgment questions, which were 96.67 % and 73.33 %, respectively. In general, the accuracy of LLM objective questions was significantly improved after LoRA fine-tuning.

(2) Experiments show that both LoRA fine-tuning and RAG can enhance the text generation ability of LLM in the field of structural demolition, but the effect of RAG improvement is more significant, and the combination of the two can further improve its accuracy. In the LLM combined with LoRA fine-tuning and RAG, Qwen2.5 has the highest average accuracy of 86.67 %, which is significantly higher than other LLMs and the average level of LLM.

(3) The intelligent generation framework of structural demolition suggestions based on multi-model collaboration can analyze the demolition methods and precautions of the structure from the specific engineering situation, and put forward the demolition suggestions that are highly consistent with the characteristics of the structure according to the relevant knowledge in the field of structural demolition. Compared with CivilGPT, the MMCF proposed in this paper can focus more on the key information of the structure, and the suggestions are more targeted.

(4) The enhanced LLM based on LoRA fine-tuning and RAG technology can not only provide suggestions for structural demolition, but also help other text generation tasks in the field of structural demolition, such as generating structural demolition safety rules and demolition plan outlines. In practical applications, engineers can write Prompt to ask questions to LLM according to other specific questions and requirements to generate corresponding answers.

**References:**

Alibaba. (2025). "Tongyi Thousand Questions." Accessed January 10,2025 <https://qianwen.aliyun.com/>

Chen, J., and Bao, Y. 2025. "A Multi-Agent Large Language Model (Llm) Framework for Code-Complying Design Automation of Concrete Structures." *Available at SSRN 5193679*. http://dx.doi.org/10.2139/ssrn.5193679

Chen, N., Lin, X., Jiang, H., and An, Y. 2024. "Automated building information modeling compliance check through a large language model combined with deep learning and ontology." *Buildings*, 14(7), 1983. http://dx.doi.org/10.3390/buildings14071983

Chiu, I.-C., and Hung, M.-W. 2025. "Finance-specific large language models: Advancing sentiment analysis and return prediction with LLaMA 2." *Pacific-Basin Finance Journal*, 90, 102632. http://dx.doi.org/10.1016/j.pacfin.2024.102632

Ding, N., Qin, Y., Yang, G., Wei, F., Yang, Z., Su, Y., Hu, S., Chen, Y., Chan, C.-M., and Chen, W. 2023. "Parameter-efficient fine-tuning of large-scale pre-trained language models." *Nature Machine Intelligence*, 5(3), 220-235. http://dx.doi.org/10.1038/s42256-023-00626-4

Du, Z., Qian, Y., Liu, X., Ding, M., Qiu, J., Yang, Z., and Tang, J. 2022."GLM: General Language Model Pretraining with Autoregressive Blank Infilling." Association for Computational Linguistics, 320-335. https://doi.org/10.18653/v1/2022.acl-long.26.

Forth, K., and Borrmann, A. 2024. "Semantic enrichment for BIM-based building energy performance simulations using semantic textual similarity and fine-tuning multilingual LLM." *Journal of Building Engineering*, 95, 110312. http://dx.doi.org/10.1016/j.jobe.2024.110312

Guo, Z., Xia, L., Yu, Y., Ao, T., and Huang, C. 2024. "Lightrag: Simple and fast retrieval-augmented generation." [In chinese] *arXiv (Cornell University)*. http://dx.doi.org/10.48550/arxiv.2410.05779

Hu, E. J., Shen, Y., Wallis, P., Allen-Zhu, Z., Li, Y., Wang, S., Wang, L., and Chen, W. 2022. "Lora: Low-rank adaptation of large language models." *ICLR*, 1(2), 3. http://dx.doi.org/10.48550/arxiv.2106.09685

Jiang, G., Ma, Z., Zhang, L., and Chen, J. 2024. "EPlus-LLM: A large language model-based computing platform for automated building energy modeling." *Applied Energy*, 367, 123431. http://dx.doi.org/10.1016/j.apenergy.2024.123431

Kim, D., Kim, T., Kim, Y., Byun, Y.-H., and Yun, T. S. 2024. "A ChatGPT-MATLAB framework for numerical modeling in geotechnical engineering applications." *Computers and geotechnics*, 169, 106237. http://dx.doi.org/10.1016/j.compgeo.2024.106237

Lee, C.-Y., and Lai, I.-W. "Enhancing Solution Diversity in Arithmetic Problems using Fine-Tuned AI Language Model." *Proc., 2024 International Conference on Consumer Electronics-Taiwan (ICCE-Taiwan)*, IEEE, 515-516. https://doi.org/10.1109/icce-taiwan62264.2024.10674425.

Lin, C.-Y. "Rouge: A package for automatic evaluation of summaries." *Proc., Text summarization branches out*, 74-81. https://aclanthology.org/W04-1013/

Liu, Z., Luo, C., Fu, D., Gui, J., Zheng, Z., Qi, L., and Guo, H. 2022. "A novel transfer learning model for traditional herbal medicine prescription generation from unstructured resources and knowledge." *Artificial Intelligence in Medicine*, 124, 102232. http://dx.doi.org/10.1016/j.artmed.2021.102232

Luo, H., Sun, Q., Xu, C., Zhao, P., Lou, J., Tao, C., Geng, X., Lin, Q., Chen, S., and Zhang, D. 2023. "Wizardmath: Empowering mathematical reasoning for large language models via reinforced

evol-instruct." *arXiv preprint arXiv:2308.09583*. http://dx.doi.org/10.48550/arxiv.2308.09583

Meng, S., Zhou, Y., Zheng, Q., Liao, B., Chang, M., Zhang, T., and Djerrad, A. 2024. " SeisGPT: A Physics-Informed Data-Driven Large Model for Real-Time Seismic Response Prediction" *arXiv preprint arXiv:2410.20186*. https://doi.org/10.48550/arXiv.2410.20186

OPENAI (2022). "ChatGPT." Accessed January 10,2025 <https://chatgpt.com/>.

Papineni, K., Roukos, S., Ward, T., and Zhu, W.-J. 2002."Bleu: a method for automatic evaluation of machine translation." *Proc., Proceedings of the 40th annual meeting of the Association for Computational Linguistics*, 311-318. https://doi.org/10.3115/1118162.1118168.

Pavlyshenko, B. M. 2023. "Financial news analytics using fine-tuned llama 2 gpt model." *arXiv preprint arXiv:2308.13032*. http://dx.doi.org/10.48550/arxiv.2308.13032

Płoszaj-Mazurek, M., and Ryńska, E. 2024. "Artificial intelligence and digital tools for assisting low-carbon architectural design: merging the use of machine learning, large language models, and building information modeling for life cycle assessment tool development." *Energies*, 17(12), 2997. http://dx.doi.org/10.3390/en17122997

Prieto, S. A., Mengiste, E. T., and García de Soto, B. 2023. "Investigating the use of ChatGPT for the scheduling of construction projects." *Buildings*, 13(4), 857. http://dx.doi.org/10.48550/arxiv.2302.02805

Pu, H., Mi, J., Lu, S., and He, J. "Rokepg: Roberta and knowledge enhancement for prescription generation of traditional chinese medicine." *Proc., 2023 IEEE International Conference on Bioinformatics and Biomedicine (BIBM)*, IEEE, 4615-4622. 10.1109/BIBM58861.2023.10385636

Pu, H., Yang, X., Li, J., and Guo, R. 2024. "AutoRepo: A general framework for multimodal LLM-based automated construction reporting." *Expert Systems with Applications*, 255, 124601. http://dx.doi.org/10.1016/j.eswa.2024.124601

Qin, S., Guan, H., Liao, W., Gu, Y., Zheng, Z., Xue, H., and Lu, X. 2024. "Intelligent design and optimization system for shear wall structures based on large language models and generative artificial intelligence." *Journal of Building Engineering*, 95, 109996. http://dx.doi.org/10.1016/j.jobe.2024.109996

Roziere, B., Gehring, J., Gloeckle, F., Sootla, S., Gat, I., Tan, X. E., Adi, Y., Liu, J., Sauvestre, R., and Remez, T. 2023. "Code llama: Open foundation models for code." *arXiv preprint arXiv:2308.12950*. http://dx.doi.org/10.48550/arxiv.2308.12950

Singhal, K., Azizi, S., Tu, T., Mahdavi, S. S., Wei, J., Chung, H. W., Scales, N., Tanwani, A., Cole-Lewis, H., and Pfohl, S. 2023. "Large language models encode clinical knowledge." *Nature*, 620(7972), 172-180. http://dx.doi.org/10.48550/arxiv.2212.13138

Uddin, S. J., Albert, A., Ovid, A., and Alsharef, A. 2023. "Leveraging ChatGPT to aid construction hazard recognition and support safety education and training." *Sustainability*, 15(9), 7121. http://dx.doi.org/10.3390/su15097121

University, Tonji. (2024). "CivilGPT." Accessed January 10,2025 <https://civilgpt.tongji.edu.cn/>.

Wang, Y., Le, H., Gotmare, A. D., Bui, N. D., Li, J., and Hoi, S. C. 2023. "Codet5+: Open code large language models for code understanding and generation." *arXiv preprint arXiv:2305.07922*. http://dx.doi.org/10.18653/v1/2023.emnlp-main.68

Xu, S., Luo, Y., and Shi, W. 2024."Geo-LLaVA: A Large Multi-Modal Model for Solving Geometry Math Problems with Meta In-Context Learning." *Proc., Proceedings of the 2nd Workshop on Large Generative Models Meet Multimodal Applications*, 11-15.

https://doi.org/10.1145/3688866.3689124.

Zhang, C., Lu, J., and Zhao, Y. 2024. "Generative pre-trained transformers (GPT)-based automated data mining for building energy management: Advantages, limitations and the future." *Energy and Built Environment*, 5(1), 143-169. http://dx.doi.org/10.1016/j.enbenv.2023.06.005

Zhang, C., Zhang, J., Zhao, Y., and Lu, J. 2024. "Automated data mining framework for building energy conservation aided by generative pre-trained transformers (GPT)." *Energy and Buildings*, 305, 113877. http://dx.doi.org/10.1016/j.enbuild.2023.113877

Zhang, H., and Gu, M. 2018. "Research and Application of BIM Model Intelligent Inspection Tool." *Civil engineering information technology*, [In Chinese] 10(02), 1-6. http://dx.doi.org/10.16670/j.cnki.cn11-5823/tu.2018.02.01

Zhang, H., and Gu, M. 2021. "The application of BIM model intelligent inspection tool in the review platform and fire inspection."*Civil engineering information technology*, [In Chinese]13(01), 1-7. http://dx.doi.org/10.16670/j.cnki.cn11-5823/tu.2021.01.01

Zhang, J., Zhang, C., Lu, J., and Zhao, Y. 2025. "Domain-specific large language models for fault diagnosis of heating, ventilation, and air conditioning systems by labeled-data-supervised fine-tuning." *Applied Energy*, 377, 124378. http://dx.doi.org/10.1016/j.apenergy.2024.124378

Zheng, Z., Chen, K.-Y., Cao, X.-Y., Lu, X.-Z., and Lin, J.-R. 2023. "Llm-funcmapper: Function identification for interpreting complex clauses in building codes via llm." *arXiv preprint arXiv:2308.08728*. http://dx.doi.org/10.48550/arxiv.2308.08728

Zhuang, Y., Yu, L., Jiang, N., and Ge, Y. 2025. "TCM-KLLaMA: Intelligent generation model for Traditional Chinese Medicine Prescriptions based on knowledge graph and large language model." *Computers in Biology and Medicine*, 189, 109887. http://dx.doi.org/10.1016/j.compbiomed.2025.109887